%% file: ijcai26.tex
\documentclass{article}
\usepackage{ijcai26}

\usepackage{times}
\usepackage{soul}
\usepackage{url}
\usepackage[hidelinks]{hyperref}
\usepackage[utf8]{inputenc}
\usepackage[small]{caption}
\usepackage{graphicx}
\usepackage{amsmath}
\usepackage{amsthm}
\usepackage{booktabs}
\usepackage{algorithm}
\usepackage{algorithmic}
\usepackage[switch]{lineno}

\usepackage{xspace}
\usepackage{subcaption}
\usepackage{multirow}
\usepackage{pifont}
\usepackage{xcolor}
\usepackage[most]{tcolorbox}
\usepackage{amsfonts}

\title{CAMERA: Adapting to Semantic Camouflage \\in Unsupervised Text-Attributed Graph Fraud Detection}

\author{
Junjun Pan$^1$
\and
Yixin Liu$^{1*}$\and
Yu Zheng$^{1*}$\and
Lianhua Chi$^2$\and
Alan Wee-Chung Liew$^1$\and
Shirui Pan$^1$
\\
\affiliations
$^1$School of Information and Communication Technology, Griffith University, Australia\\
$^2$Department of Computer Science and Information Technology, La Trobe University, Australia\\
\emails
junjun.pan@griffithuni.edu.au, 
\{yixin.liu, yu.zheng, a.liew, s.pan\}@griffith.edu.au, 
l.chi@latrobe.edu.au
}

\newcommand{\ourmethod}{CAMERA\xspace}
\begin{document}

\maketitle
\renewcommand{\thefootnote}{\fnsymbol{footnote}}
\footnotetext[1]{Corresponding authors.}

\begin{abstract}
Text-attributed graph fraud detection (TAGFD) plays a critical role in preventing fraudulent activities on online social and e-commerce platforms. However, to evade detection, fraudsters continuously evolve their camouflaging strategies by deliberately mimicking textual responses of benign users, thereby concealing their malicious purposes. This phenomenon, referred to as \textbf{semantic camouflage}, fundamentally undermines commonly relied assumptions on how structural and attribute cues can be exploited to identify fraudsters, and makes it difficult to spot fraudsters with unsupervised TAGFD. To bridge the gaps, we propose a \textbf{C}ase-\textbf{A}daptive \textbf{M}ulti-cue \textbf{E}xpert f\textbf{RA}mework (CAMERA) for unsupervised TAGFD. 
CAMERA employs an ego-decoupled mixture-of-experts architecture, where each expert specializes in modeling a distinct type of fraud-indicative cue. A context-informed gating model is introduced to jointly consider the ego node representation and its local neighborhood context for adaptive integration of cues learned by different experts. Furthermore, CAMERA leverages the inherent rarity of fraudsters to support unsupervised one-class learning with expert-level objectives that encourage modeling dominant benign patterns, thereby enabling reliable unsupervised detection of camouflaged fraudsters. 
Experiments on 4 challenging datasets show that CAMERA consistently outperforms competitors, showing its effectiveness against semantically camouflaged fraudsters. Code available at https://github.com/CampanulaBells/CAMERA

\end{abstract}

\section{Introduction}
\input{Chapters/Intro}

\section{Related Work}
\input{Chapters/Related_Work}

\section{Preliminary}
\input{Chapters/Preliminary}

\section{Methodology}
\input{Chapters/Method}

\section{Experiments}

\input{Chapters/Experiment}

\section{Conclusion}
In this paper, we propose a novel unsupervised TAGFD framework, \ourmethod, that enables detection against evolved fraudsters that engage in semantic camouflage to mimic benign behaviors and evade detection.
By utilizing ego-decoupled MoE architecture together with a one-class unsupervised training objective, \ourmethod can adaptively integrate different fraud-indicative cues to address evolving camouflaging strategies. 
Comprehensive empirical results show the effectiveness and robustness of \ourmethod in unsupervised TAGFD, validating \ourmethod as a practical solution for detecting evolved fraudsters in real-world scenarios.



\section*{Acknowledgements}
The work of S. Pan was partially supported by the Australian Research Council (ARC) under Grant Nos. DP240101547 and FT210100097. The work of Y. Liu was partially supported by the ARC under Grant No. DE260101172. 

\bibliographystyle{named}
\bibliography{ijcai26}
\clearpage
\appendix
\input{Appendix_Content}

\end{document}

%% file: Chapters/Intro.tex
With the continued advancement of information technology, graph-structured data has become ubiquitous in online services such as e-commerce~\cite{yu2023group} and social media~\cite{hu2023cost}. This ubiquity has been accompanied by increasing malicious activities, including fake reviews and spam messages, which underscores the importance of graph fraud detection (GFD)~\cite{pan2025survey,cai2025out}. Since many real-world platforms are inherently associated with textual content, text-attributed graph fraud detection (TAGFD) has emerged as an essential research direction for identifying suspicious behaviors embedded in text-attributed graphs~\cite{yang2025flag,qian2026dynhd}. By jointly capturing network structures and attribute information, TAGFD plays a vital role in safeguarding the online platforms~\cite{pan2026explainable}. 

\begin{figure}[t]
    \centering
    \begin{subfigure}[b]{0.95\linewidth}
        \centering
        \includegraphics[width=0.95\linewidth]{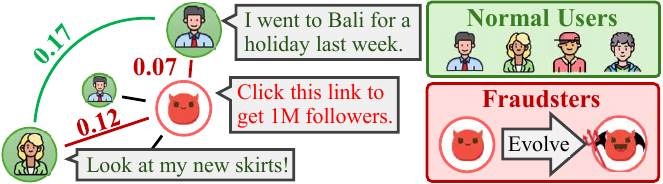}
        \caption{Early fraudsters hide in benign communities.}
        \label{fig:a}
    \end{subfigure}

    \begin{subfigure}[b]{0.95\linewidth}
        \centering
        \includegraphics[width=0.95\linewidth]{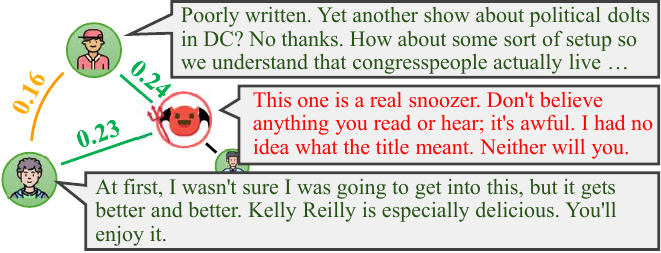}
        \caption{Evolved fraudsters employ semantic camouflage.}
        \label{fig:b}
    \end{subfigure}


    \caption{Illustration of fraudster evolution in text-attributed graphs, where the edge weights indicate the affinities between two nodes.} 
    \label{fig:intro_1}
\end{figure}

Despite the remarkable progress of detection techniques in recent years, fraudsters also continuously adapt their camouflage strategies to evade detection~\cite{yang2025grad}. As illustrated in Figure~\ref{fig:intro_1}, earlier fraudsters primarily operated at the structure level, where fraudulent nodes attempted to blend into the benign community by connecting themselves to benign users~\cite{dou2020enhancing}. While existing fraud detection methods could still handle these topology-level camouflages by pruning heterophily edges~\cite{qiao2023truncated}, evolved fraudsters also deliberately hide their malicious intent by mimicking normal characteristics, which significantly increases the difficulty of detection. 
Going beyond structure-level blending, these evolved fraudsters further engage in 
\textbf{semantic camouflage}, i.e., they reshape their language to mimic benign expressions like criticism, while quietly advancing malicious goals like review manipulation or the spread of misleading information.

To counter this semantic camouflage, recent TAGFD studies have explored supervised approaches to learn semantic-level anomalies-indicative cues. For instance, FLAG~\cite{yang2025flag} leverages a large language model (LLM) to extract both shared and discriminative contextual signals from collections of fraudulent and normal texts, while DGP~\cite{li2025dgp} integrates graph structure into LLM fine-tuning. However, these methods typically rely on labels to find the cue from semantics. In real-world applications, however, obtaining reliable annotations for GFD is often costly and impractical. This limitation motivates us to explore a more practical research problem, i.e., \textbf{{unsupervised TAGFD}}, where fraudsters are identified without access to ground-truth labels.

\begin{figure}[t]
    \centering
    \includegraphics[width=\linewidth]{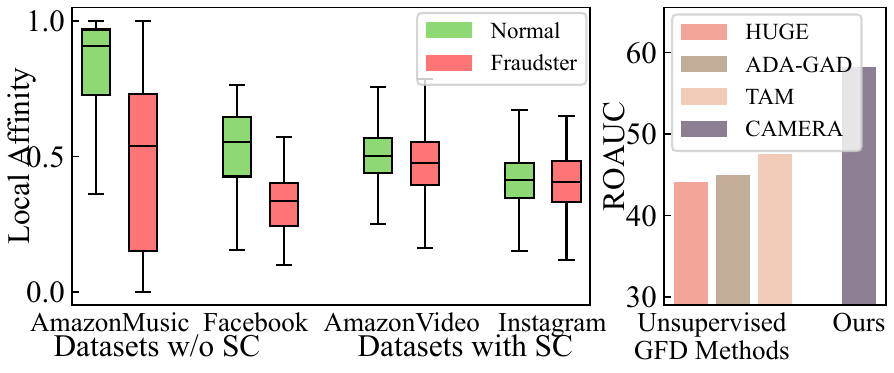}
    \caption{Left: Local affinity of different datasets with and without semantic camouflage (SC). Right: ROAUC on Instagram dataset. }
    \label{fig:intro_2}
\end{figure}

To handle this new research problem, a naive solution is to extend existing unsupervised GFD methods~\cite{qiao2023truncated,he2024ada,pan2025label} by incorporating a text encoder to extract textual representation as graph features. 
These methods assume that fraudsters disrupt local affinity~\cite{huang2023unsupervised}, and hence the structure and attribute information are combined to compute affinity as fraud-indicative cues, either directly detecting fraudsters~\cite{pan2025label} or revealing camouflage through edge pruning~\cite{qiao2023truncated,he2024ada}. 
However, as shown in Figure~\ref{fig:intro_2}, fraudsters with semantic camouflage induce only minor changes in local affinity. As a result, simply extending unsupervised GFD methods to TAGFD scenarios with semantic camouflage can lead to degraded performance. In light of this, a core question arises: 

\vspace{-1mm}
\begin{tcolorbox}[
  colback=gray!6,
  colframe=black!40,
  boxrule=0.6pt,
  arc=2mm,
  left=1.2mm,right=1.2mm,top=1mm,bottom=1mm
]
\textit{How can we perform unsupervised TAGFD under semantic camouflage without access to ground-truth labels?}
\end{tcolorbox}
\vspace{-1mm}

In answering this question, we identify two critical challenges: 
\noindent \textit{\textbf{Challenge 1} - Adaptive integration of fraud-indicative cues.} 
Effective fraud detection relies on capturing diverse fraud-indicative cues from both graph structure and node attributes. However, evolving camouflaging strategies and increasingly complex real-world GFD scenarios continually invalidate the predetermined assumptions regarding the integration of these cues. This motivates the need for approaches that adaptively integrate multiple fraud-indicative cues to detect fraudsters with diverse camouflaging strategies. 
\noindent \textit{\textbf{Challenge 2} - Capturing fraud-indicative cues under unsupervised settings.} 
Evolved fraudsters not only hide in benign communities but also actively camouflage in the semantic domain, making their malicious evidence hard to recognize. Such semantic camouflage undermines many commonly adopted heuristic designs, making it particularly challenging to highlight minor deviations from normal patterns without human-annotated labels. 
As a result, how to leverage the minority nature of fraudsters becomes critical for building an effective TAGFD model.

To fill the gap, we propose a \textbf{C}ase-\textbf{A}daptive \textbf{M}ulti-cue \textbf{E}xpert f\textbf{RA}mework (CAMERA) for unsupervised TAGFD against evolved fraudsters.  
To tackle \textbf{\textit{Challenge 1}}, \ourmethod introduces an ego-decoupled Mixture-of-Experts (MoE) architecture that allows adaptive integration of multiple fraud-indicative cues, where each expert specializes in a distinct type of anomaly signal. By explicitly decoupling the specialization of each expert, the experts can extract complementary fraud-indicative cues that highlight diverse fraud patterns. 
Moreover, we further enhance the gating model with local context to guide the adaptive integration in a finer manner. 
To address \textbf{\textit{Challenge 2}}, \ourmethod leverages the inherent rarity of fraudsters to support unsupervised training. Through one-class learning combined with expert-specific loss, each expert is encouraged to model dominant benign patterns, making malicious deviations introduced by fraudsters more pronounced. Based on these deviation-aware representations, a lightweight parameter-free fraud detector identifies nodes that deviate from normal patterns, enabling unsupervised detector training. Together, \ourmethod avoids reliance on rigid assumptions about fraud behaviors while integrating multiple fraud-indicative cues, providing a solid solution for unsupervised TAGFD in the presence of semantically camouflaged fraudsters. In summary, our contributions are threefold: 

\noindent \textbf{Problem.} To the best of our knowledge, we are the first to formally address the challenge of unsupervised TAGFD under the semantic camouflage scenarios, where fraudsters can mimic benign behaviors to hide their malicious intent. 

\noindent \textbf{Method.} We propose \ourmethod, a novel unsupervised TAGFD framework that employs a MoE architecture to adaptively integrate multiple fraud-indicative cues to reveal semantically camouflaged fraudsters without supervision.

\noindent \textbf{Experiments.} We conduct extensive experiments to demonstrate the superior performance of \ourmethod over the state-of-the-art methods on four real-world TAGFD datasets under unsupervised learning scenarios.

%% file: Chapters/Related_Work.tex
In this section, we provide a summary of two related areas. Detailed reviews are provided in Appendix~A. 

\noindent\textbf{Fraud Detection on Attributed Graph.} 
Early studies formulate graph fraud detection (GFD) as a supervised class-imbalanced classification problem, relying on labeled fraud instances and sampling-based techniques~\cite{liu2021pick}. However, real-world fraudsters often camouflage themselves by blending into benign communities via heterophilic edges~\cite{gao2023addressing}, which substantially degrades supervised models. To address this, later works explicitly incorporate this finding into model design~\cite{zhao2025freegad}, for example, by pruning heterophilic edges~\cite{gao2023addressing} or mitigating representation shift caused by fraudsters~\cite{tang2022rethinking}. However, the cost of obtaining annotations can limit their application in diverse real-world scenarios. To address this, follow-up unsupervised GFD studies incorporate heuristic assumptions to obtain heterophilic edges. For example, TAM~\cite{qiao2023truncated} and ADA-GAD~\cite{he2024ada}  iteratively compute pseudo-labels to prune heterophilic edges, while HUGE~\cite{pan2025label} establishes a label-free heterophily measure as guidance, thereby training the fraud detector through alignment loss. While recent works further enhance the generalizability by transfer to unseen domains~\cite{pan2026correcting,zhao2026fedcigar} or building one-for-all GAD models~\cite{liu2026few}, they ignore the rich semantics contained in text attributes, thereby limiting their application.

\noindent\textbf{Fraud Detection on Text-Attributed Graph}
Many real-world graphs contain rich textual context, making text-attributed GFD (TAGFD) an important research direction, where evolved fraudsters employ semantic camouflage to escape detection. 
As a representative method, CoLL~\cite{xu2025court} incorporates LLMs to generate anomaly evidence, but as evolved fraudsters deliberately mimic benign user behaviors, yet when fraudsters deliberately mimic benign behaviors, LLMs are easily misled, resulting in incorrect or misleading evidence.  Another line of work integrates structural and textual information into unified graph contrastive learning frameworks~\cite{liu2025towards,xu2025text}. Nevertheless, semantic camouflage typically induces only subtle ego-neighbor divergence compared to injected structural or feature anomalies~\cite{ding2019deep}, weakening the effectiveness of contrastive objectives. These limitations motivate us to address semantic camouflage in unsupervised TAGFD.

%% file: Chapters/Preliminary.tex

\noindent\textbf{Notations.} Let $\mathcal{G} = (\mathcal{V}, \mathcal{E}, \mathcal{T})$ represents a text-attributed graph, where $\mathcal{V}$ is the set of nodes and $\mathcal{E}$ is the set of edges. We denote the number of nodes and edges as $n$ and $m$, respectively. $\mathcal{T} = \{ t_i \mid v_i \in \mathcal{V} \}$ denotes the set of textual attributes, where $t_i \in \mathcal{D}^{L_i}$ is the text sequence associated with node $v_i$, $\mathcal{D}$ represents the dictionary of words, and $L_i$ is the length of the text sequence. The graph structure can also be represented as adjacent matrix $\mathbf{A} \in \{0,1\}^{n \times n}$, with $\mathbf{A}_{i,j} = 1$ if there exists an edge between nodes $v_i$ and $v_j$, and $\mathbf{A}_{i,j} = 0$ otherwise. We denote the set of neighbors of the $i$ th node as $\mathcal{N}(v_i) = \{ v_j \mid \mathbf{A}_{i,j} = 1 \}$.

\noindent\textbf{Problem Definition.} In the context of TAGFD, each node $v_i \in \mathcal{V}$ has a label $y_i \in \{0, 1\}$, where $y_i = 0$ indicates a benign node and $y_i = 1$ represents a fraudulent node. A commonly accepted assumption is that the number of benign nodes is significantly greater than the number of fraudulent nodes. In unsupervised scenarios, labels are unavailable during the training stage. Given a text-attributed graph $\mathcal{G}$, the goal of unsupervised TAGFD is to learn an scoring function $f:\mathcal{V} \xrightarrow{} \mathcal{R}$ to identify whether a node $v_i$ is suspicious by predicting a fraud score $s_i^{\text{fraud}}=f(v_i, \mathcal{G})$, where a higher score indicates a greater likelihood that the node is fraudulent.

%% file: Chapters/Method.tex
\begin{figure}
\centering
\includegraphics[width=.98\linewidth]{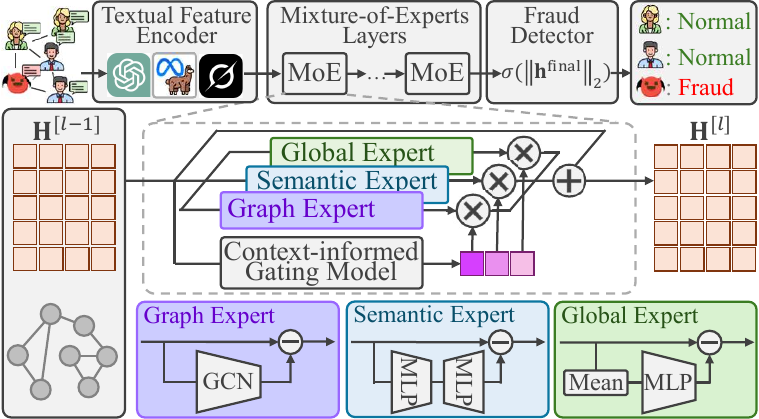}
\caption{
Overall framework of \ourmethod. }%
%
\label{fig:architecture}
\end{figure}

In this section, we provide an overview of \ourmethod. As shown in Figure~\ref{fig:architecture}, we first encode textual node attributes using an LLM to extract high-quality node embeddings. Building on these embeddings, the ego-decoupled Mixture-of-Experts (MoE) layers extract complementary deviation signals, with each expert specializing in a distinct type of fraud-indicative cue. As what constitutes anomalous behavior differs across communities, the contributions of the MoE experts are weighted by a context-informed gating model that leverages both ego and neighborhood embeddings as priors. By learning dominant normal patterns in an unsupervised manner to expose subtle deviations, \ourmethod enables adaptive integration of fraud-indicative cues for detecting camouflaged fraudsters without access to ground-truth labels.

\subsection{Textual Feature Encoder}
Conventional GFD methods often rely on pre-extracted shallow features (e.g., TF-IDF), which are insufficient for capturing the nuanced evidence for malicious purposes required to detect fraud under semantic camouflage. To address these limitations, we employ a pre-trained LLM to encode the textual attributes, providing richer semantic representations for the identification of camouflaged fraudsters. Specifically, given the textual attributes  $t_i$ of node $v_i$, we adopt an LLM-based feature encoder to transform the input text into a dense semantic representation $\mathbf{x}_i$  as the node attribute: $\mathbf{x}_i = \text{LLM}(t_i)$. The representations are subsequently used as the input to the downstream MoE modules, i.e., $\mathbf{H}^{[0]} = \left[\mathbf{x}_1,\dots,\mathbf{x}_{n}\right]\in \mathbb{R}^{n \times d}$, where $d$ is the dimension of LLM-generated representations and each row $\mathbf{h}^{[0]}_i$ is the representation vector of node $v_i$. Compared with small-size text encoders such as SentenceBERT~\cite{reimers2019sentence}, which were utilized by previous studies of text-attributed graphs, the LLM-based encoder exhibits a stronger semantic compression capability. This is particularly critical under unsupervised TAGFD settings, where subtle cues must be preserved without supervision signals.

\subsection{Ego-decoupled MoE Architecture}

While the pre-trained LLM-based encoder captures deep semantic information, spotting evolving camouflaged fraudsters requires the effective and adaptive integration of other fraud-indicative cues, ranging from local structural patterns~\cite{pan2025label} to global semantic distributions~\cite{jin2021anemone}. To address this challenge, we adopt a MoE architecture~\cite{li2026towards}, where each expert is highly specialized and capable of capturing one type of discriminative fraud-indicative cues, providing strong and complementary representations for effective detection against camouflaged fraud~\cite{tan2025bisecle}. To model the interactions among heterogeneous cues, we further employ a multi-layer MoE design, where each MoE layer adaptively routes node representations to different experts and refines the fused representation, which hierarchically integrates the fraud-indicative cues. Moreover, a gating network complements the experts by dynamically learning to weight their contributions for each node. This adaptive mechanism allows the model to flexibly integrate fraud-indicative cues without relying on dataset-specific heuristics or domain knowledge, ensuring the most informative cues are prioritized.

\subsubsection{Ego-Decoupled MoE Layer}

As the essential building block of \ourmethod, we first introduce the proposed MoE layer, which adaptively routes and integrates heterogeneous fraud-indicative cues. 
Specifically, taking the $l$-th layer as an example, a standard MoE layer can be written as:

\begin{equation}
\mathbf{H}^{[l]} = \sum_{k} \mathrm{diag}\!\big(g_k^{[l]}(\mathbf{H}^{[l-1]}, \mathbf{A})\big)\,
e_k^{[l]}(\mathbf{H}^{[l-1]}, \mathbf{A}).
\end{equation}

\noindent where $\mathbf{H}^{[l]} \in \mathbb{R}^{n \times d}$ denotes the output embedding matrix, $e_k^{[l]}(\mathbf{H}^{[l-1]}, \mathbf{A})\in \mathbb{R}^{n \times d}$ is the embedding matrix produced by the expert $k$, and $g_k^{[l]}(\mathbf{H}^{[l-1]}, \mathbf{A})\in \mathbb{R}^{n}$ denotes the corresponding gating weight vector.


Despite its effectiveness in modeling heterogeneous fraud-indicative cues, this generic formulation does not explicitly enforce functional decoupling among experts. Similar to the over-smoothing effect in graph neural networks~\cite{rusch2023survey}, redundant expert functions can produce overly homogeneous representations~\cite{liu2023diversifying}, which may obscure critical malicious cues. To address this issue, we propose an ego-decoupled MoE layer that separates expert-specific deviation signals from shared ego features:
\begin{equation}
\mathbf{H}^{[l]} = \mathbf{H}^{[l-1]} + \sum_{k} \mathrm{diag}\!\big(g_k^{[l]}(\mathbf{H}^{[l-1]}, \mathbf{A}) \big) e_k^{[l]}(\mathbf{H}^{[l-1]}, \mathbf{A}).
\end{equation} 
By isolating shared ego embeddings, each expert focuses on complementary fraud signals rather than redundant information. Additionally, the skip connection integrates multi-hop graph features across layers, further enhancing the inter-layer information communication within the MoE model for fraud detection~\cite{dong2025spacegnn}.

\subsubsection{Expert Specialization} 

MoE architectures typically achieve expert functional divergence through large-scale unsupervised pretraining~\cite{fedus2022switch} or domain partitioning~\cite{gururangan2022demix} with supervision. However, in unsupervised TAGFD, neither human annotations nor sufficiently large curated datasets are available, which hinders the operationally identical experts from learning diverse and discriminative fraud-indicative cues. Therefore, to ensure functionality divergence, we explicitly design the operation of each expert $e_k^{[l]}$ to specialize in a distinct fraud-indicative cue, ensuring functional differentiation and capturing complementary cues critical for detecting camouflaged fraudsters. Concretely, three specialized experts, i.e., the graph expert $e_\text{graph}^{[l]}$, the semantic expert, and the global expert, are instantiated in \ourmethod, which are defined as follows.

\noindent \textbf{Graph Expert.} The graph expert focuses on capturing \textit{structural deviation} signals, thereby spotting any unusual structural patterns that cannot be encoded with text attributes alone. We first encode local structural patterns using a GCN layer and compute the residual between ego representations and aggregated neighborhood features to expose anomalous structural discrepancies. Specifically, for each node $v_i$ with representation $\mathbf{h}^{[l-1]}_i$, the operation of the graph expert can be written as:
\begin{equation}
e_\text{graph}^{[l]}(\mathbf{h}^{[l-1]}_i, \mathbf{A})
=
\mathbf{h}_i^{[l-1]}
-
\operatorname{GCN}\!\left(\mathbf{h}_i^{[l-1]}, \{\mathbf{h}_j^{[l-1]}\}_{v_j\in\mathcal{N}(v_i)}\right).
\end{equation}

\noindent \textbf{Semantic Expert.} To precisely spot any malicious semantic cues, the semantic expert employs an MLP-based autoencoder to learn a compact representation of benign semantics, and uses the difference between that and the input embedding to reveal \textit{semantic deviations}:
\begin{equation}
e_\text{semantic}^l(\mathbf{h}_i^{[l-1]}) = \mathbf{h}_i^{[l-1]} - \text{Decoder}(\text{Encoder}(\mathbf{h}_i^{[l-1]})).
\end{equation}
Hence, the semantic deviation encodes fine-grained malicious cues of the text, which is critical for detecting semantically camouflaged fraudsters that mimic normal behaviors. 

\noindent \textbf{Global Expert.} While the graph and semantic experts focus on local deviations, fraudsters are inherently rare instances that deviate from the overall data distribution, which can be exposed from a global perspective. In light of this, we utilized the global expert to measure each node’s discrepancy from the dominant benign prototype:
\begin{equation}
e_\text{global}^{[l]}(\mathbf{h}_i^{[l-1]}) = \mathbf{h}_i^{[l-1]} - \text{MLP}(\mathbf{h}_\text{global}^{[l-1]}), 
\end{equation}
where $\mathbf{h}_\text{global}^{[l-1]} = \frac{1}{n} \sum_{v_j \in \mathcal{V}} \mathbf{h}_j^{[l-1]}$.
The estimated \textit{global deviation} characterizes how much a node diverges from the majority, complementing the local structural and semantic deviations captured by other experts.

Collectively, the graph, semantic, and global experts maintain functional differentiation while capturing complementary fraud residuals across structural, textual, and distributional fraud-indicative cues, which ensures that the MoE layer learns diverse and complementary fraud-indicative cues, thereby laying a solid foundation for the gating network to perform adaptive, instance-specific embedding integration.

\subsubsection{Context-informed Gating Model}
After embedding extraction by specialized experts, a gating model is responsible for dynamically weighting their contributions for every node, enabling adaptive integration of fraud-indicative cues without relying on prior assumptions or domain-specific knowledge. In standard MoE architectures, gating is typically implemented using input-dependent mechanisms such as self-attention~\cite{lewis2021base}. However, in TAGFD, the importance of different fraud-indicative cues is not solely determined by a node’s attributes, but is highly dependent on its local context. For example, in topic-focused communities, abnormal connection patterns often provide stronger evidence of fraud, whereas in more diverse communities, semantic irregularities such as misleading or toxic content become more informative. Relying only on ego features may therefore lead to suboptimal expert selection. Motivated by this observation, we propose a context-informed gating model that jointly considers the ego node representation and its local neighborhood context when determining expert contributions to allow for more fine-grained adaptation.

Specifically, to obtain the local neighborhood context $\mathbf{c}_i^{[l-1]}$ of node $v_i$, we aggregate the embeddings of its neighbors as:
\begin{equation}
\mathbf{c}_i^{[l-1]} = \frac{1}{\text{deg}(v_i)} \sum_{v_j \in \mathcal{N}(v_i)} \mathbf{h}_j^{[l-1]},
\end{equation}
where $\mathbf{h}_j^{[l-1]}$ denotes the input embedding of node $v_j$ at the $l-1$-th layer, $\mathcal{N}(v_i)$ is the neighbor set of $v_i$, and $\text{deg}(v_i)$ is its degree. The gating network then leverages both the ego representation and the local context to compute expert weights:
\begin{equation}
\begin{aligned}
 \mathbf{g}_i^{[l]} &= \text{Softmax}\big(\text{Linear}([\mathbf{h}_i^{[l-1]} \, || \, \mathbf{c}_i^{[l-1]}])\big) \in \mathbb{R}^{3}, \\
& g_k^{[l]}(\mathbf{H}^{[l-1]}, \mathbf{A}) = \{ \mathbf{g}_{i,k}^{[l]} \mid v_i \in \mathcal{V} \},
\end{aligned}
\end{equation}
where $\mathbf{g}_{i,k}^{[l]}$ represents the $k$-th entry of $\mathbf{g}_i^{[l]}$ with $k=\{1,2,3\}$ corresponds to graph, semantic, and global experts, respectively, $||$ denotes concatenation and $\text{Linear}(\cdot)$ is a learnable linear transformation. By incorporating both ego features and neighborhood context, the proposed gating model enables more informed selection and aggregation of the malicious cues captured by different experts. Together with the unsupervised training objectives, our context-aware weighting allows the MoE layer to adaptively integrate fraud-indicative cues, thereby providing the flexibility needed to effectively detect camouflaged fraudsters under the diverse and evolving conditions of TAGFD.

To encourage context-dependent expert usage and sparse gating weights, we apply an entropy-based regularization loss on the gating weights $\mathbf{g}_i$ to train the gating layer:
\begin{equation}
\mathcal{L}_\text{gating} = \sum_{l}  \sum_{k \in \{\text{graph, semantic, global}\}} - \frac{1}{N} \sum_{i=1}^N  g_{i,k}^{[l]} \log (g_{i,k}^{[l]} + \epsilon),
\end{equation}
where  $\epsilon$ is a small constant for numerical stability. Notably, during training, we block the gradient from propagating to earlier layers to ensure that the gating loss only updates the gating network, preventing it from directly updating weights of experts to ensure their specialization. By minimizing $\mathcal{L}_\text{gating}$, we encourage the gating distribution to have a sharper distribution, thereby encouraging each node to strategically utilize the relevant experts given its local context.

\subsection{Rarity-driven Unsupervised Fraud Detection}
While the proposed MoE layers are designed to adaptively integrate fraud-indicative cues, training them for TAGFD in a fully unsupervised setting remains challenging. As evolved fraudsters deliberately camouflage their malicious purpose by mimicking benign semantics, the assumptions underlying traditional detection methods (e.g., affinity assumption~\cite{qiao2023truncated}) no longer hold, thereby rendering them ineffective in handling evolved fraudsters. 

To seek alternative supervision signals, the intrinsic rarity of fraudsters in real-world datasets is a stable and persistent property that can serve as a reliable training prior. Concretely, the ``rarity assumption'' is that fraudulent behaviors can be characterized as deviations from dominant benign patterns, and hence fraudsters can be identified as low-density outliers in the learned normality space. Motivated by this insight, we draw inspiration from one-class (OC) classification to model the dominant patterns of benign nodes with a parameter-free OC fraud detector, thereby exposing the deviations introduced by minority fraudulent nodes. Meanwhile, we design an expert loss that encourages each expert to capture normal patterns, thereby highlighting any cues of fraudsters in the residual deviations. 

\subsubsection{Parameter-free One-class Fraud Detector}
Building on the discriminative embeddings produced by the MoE experts, we employ a parameter-free OC classifier that models normal patterns and identifies deviations. Specifically, the $\ell_2$ norm of each embedding serves as a normality measure, which is converted into a bounded fraud score via a sigmoid function $\sigma(\cdot)$, i.e.,
\begin{equation}
s_i = \sigma(\|\mathbf{h}^{\text{final}}_i\|_2),
\end{equation}
where $\mathbf{h}^{\text{final}}_i$ is the representation of $v_i$ at the final MoE layer. This design provides a lightweight yet effective mechanism to fully leverage the fraud-indicative cues captured by the MoE layers, while also enabling unsupervised detection of subtle fraudulent behaviors with an OC detector.

Under the assumption that fraudsters are a minority, we encourage most fraud scores to approach zero by minimizing
\begin{equation}
\mathcal{L}_\text{OC} = \frac{1}{N} \sum_{i=1}^N \text{BCE}(s_i, 0),
\end{equation}

\noindent where $\text{BCE}(\cdot,\cdot)$ denotes binary cross-entropy. Optimizing $\mathcal{L}_\text{OC}$ pushes the embeddings of normal nodes into a compact hypersphere around the origin while naturally allowing minority fraudsters to stand out due to their larger norms~\cite{wang2021one}. Leveraging the rarity assumption, the OC loss suppresses fraud scores for the majority of nodes while keeping the detector lightweight and easy to optimize. More discussion on the property of OC loss is given in Appendix B. 

\input{Tables/MainResults}

\subsubsection{Expert Loss}
Although the OC loss can supervise the whole network at a global level, the functionality of each expert could be unconstrained, which leads to expert redundancy and weak specialization. To further regularize the specialization of experts, we introduce an expert loss to enforce functional usefulness at a fine-grained expert level. 
As we explicitly design each expert to specialize in a distinct type of fraud-indicative cue, minimizing their corresponding deviation signals can naturally serve as the optimization objective. Since the majority of nodes are normal, this encourages each expert to model the patterns of benign nodes, causing the residual deviations of minority fraudulent nodes to become more pronounced. To maintain expert disentanglement, each loss is applied only to the parameters of the corresponding expert, and gradients are blocked from propagating to earlier layers or the gating network. Formally, the expert loss is computed as the average squared deviation across nodes, experts, and layers:
\begin{equation}
\mathcal{L}_\text{expert} = \sum_{l} \sum_{k \in \{\text{graph, semantic, global}\}} \frac{1}{N} \sum_{i=1}^N \| e_k^{[l]}(\mathbf{h}_i^{[l-1]}, \mathbf{A}) \|_2^2,
\end{equation}
where $N$ is the number of nodes, and $e_k^{[l]}(\cdot)$ denotes the residual deviation captured by the $k$-th expert at layer $l$. By modeling benign patterns, the experts naturally amplify the subtle malicious signals caused by camouflaged fraudsters, thereby producing discriminative embeddings $\mathbf{h}^{\text{final}}_i$ for the subsequent detector.

The overall training objective is defined by combining $\mathcal{L}_\text{OC}$ with $\mathcal{L}_\text{expert}$ and $\mathcal{L}_\text{gating}$, which is written as:
\begin{equation}
\mathcal{L} = \mathcal{L}_\text{expert} + \alpha \mathcal{L}_\text{gating} + \beta \mathcal{L}_\text{OC},
\end{equation}
where $\alpha$ and $\beta$ are trade-off hyperparameters. The overall algorithm and complexity analysis of \ourmethod is given in Appendices C and D, respectively.

%% file: Tables/MainResults.tex
\begin{table*}[t]

\centering
\resizebox{\textwidth}{!}{
\begin{tabular}{c|l|cccc|cccc}
\toprule
\multirow{2}{*}{Type} & \multirow{2}{*}{Method}
 & \multicolumn{4}{c|}{AUROC (\%)} 
 & \multicolumn{4}{c}{AUPRC (\%)} \\
 &  
 & Reddit & Instagram  & AmazonVideo & YelpChi 
 & Reddit & Instagram  & AmazonVideo & YelpChi \\
\midrule
\multirow{3}{*}{GAD}
 & DOMINANT (SDM'19)  &  62.89±0.09 &  45.35±0.08 & OOM & OOM & 20.60±0.04 & 8.92±0.01  & OOM & OOM \\
  & CoLA (TNNLS'21)     & 58.63±1.81 &  45.05±0.20 & 51.06±1.95 &  48.62±0.16
 & 13.44±0.60 & 8.95±0.06  & 12.83±0.56  & 14.08±0.08 \\
 & PREM (ICDM'23) & 54.85±4.99 &  54.38±3.59 & 58.08±1.91 & OOM & 12.16±2.31 & 11.51±1.16 &  15.06±0.68 &  OOM\\
\midrule
\multirow{3}{*}{GFD}
 & TAM (NeurIPS'23)  & 63.07±0.10 & 47.51±0.06 & OOM & OOM & 14.76±0.11 & 10.40±0.07 & OOM &  OOM\\
 & ADA-GAD (AAAI'24) & 64.09±0.02 & 44.98±0.01 & OOM & OOM & \textbf{21.78±0.02} & 8.87±0.01 & OOM & OOM \\
 & HUGE (AAAI'25) & 60.64±0.78 & 44.07±0.17 &  51.77±0.55 & OOM & 12.80±0.57 & 9.00±0.15 & 12.49±0.17 &  OOM \\
\midrule
\multirow{3}{*}{TAGAD}
 & TAGAD (arXiv'25) & 56.65±0.12 & 44.83±0.06 & 51.55±0.12 & 42.21±0.11 & 11.64±0.04 & 9.04±0.02 & 12.14±0.04 &  10.78±0.02  \\
 & CMUCL (ECAI'25) & 60.72±0.15 & 44.41±0.30 & 53.99±0.22 & OOM & 13.52±0.14 &  8.81±0.11  & 13.80±0.19  &  OOM\\
 & CoLL (MM'25) & 59.26±0.06 & 44.02±0.72 & 50.27±1.26 &  47.25±0.32 & 13.22±0.18 & 8.88±0.20 & 11.99±0.41 & 12.50±0.09 \\
\midrule
\multirow{1}{*}{TAGFD}
 &  \ourmethod (Ours) &\textbf{ 65.09±0.04} & \textbf{58.21±0.29} & \textbf{63.05±0.60} &\textbf{ 61.74±0.04} & 18.99±0.04 & \textbf{14.23±0.54} &\textbf{ 17.37±0.83} & \textbf{18.47±0.02}\\
\bottomrule
\end{tabular}
}
\caption{Performance comparison in terms of AUROC and AUPRC, with best results in \textbf{bold}. OOM means out-of-memory on a 24GB GPU.}

\label{table:MainResults}

\end{table*}

%% file: Chapters/Experiment.tex
\subsection{Experimental Setup}
\noindent \textbf{Datasets. }We conduct experiments on four public GAD datasets spanning diverse application domains, including Reddit~\cite{li2024glbench}, Instagram~\cite{li2024glbench}, AmazonVideo~\cite{mcauley2013amateurs}, and YelpChi~\cite{rayana2015collective}. Following the previous work~\cite{li2025dgp,yang2025flag}, we construct the text-attributed graph for fraud detection purposes, and convert all graphs to homogeneous undirected graphs for our experiments. Detailed dataset statistics are summarized in Appendix E.

\noindent \textbf{Baseline and Evaluation Metrics. }We compare our approach against 9 state-of-the-art methods. CoLA~\cite{liu2021anomaly}, DOMINANT~\cite{ding2019deep}, and PERM~\cite{pan2023prem} represent contrastive learning, reconstruction-based, and affinity-based paradigms for unsupervised graph anomaly detection (GAD). TAM~\cite{qiao2023truncated}, ADA-GAD~\cite{he2024ada}, and HUGE~\cite{pan2025label} focus on unsupervised GFD by pruning heterophily edges to mitigate the impact of camouflaged fraudsters. We also include recent studies that incorporate text attributes (TAGAD), namely TAGAD~\cite{liu2025towards}, CMUCL~\cite{xu2025text}, and CoLL~\cite{xu2025court}. To ensure a fair comparison, we use OpenAI’s text-embedding-3-small model to encode textual attributes for \ourmethod and baselines. We use gpt-4o-mini for TAGAD methods that incorporate LLM. 
The area under the receiver operating characteristic curve (AUROC) and the area under the precision-recall curve (AUPRC) are used as the evaluation metrics. Average results with standard deviations are reported over five runs with different random seeds. The implementation details and more experiments are presented in Appendices E and F, respectively. 
\input{Tables/Ablation_Feature}

\subsection{Experimental Results}
\subsubsection{Performance Comparison}
The comparison results of \ourmethod is illustrated in Table~\ref{table:MainResults}. 
From the table, we make the following key observations: \ding{182}~\ourmethod outperforms SOTA methods in most datasets, with the only exception being AUPRC on Reddit. These results demonstrate the effectiveness of~\ourmethod in diverse real-world scenarios. \ding{183}~Among the baselines, methods that address structure camouflage by pruning heterophilic edges (TAM, ADA-GAD, HUGE) achieve strong performance on Reddit but perform poorly on other datasets such as Instagram, revealing the limitation of static assumptions that fail to capture evolving fraudster characteristics. In contrast, the adaptive fraud-indicative cues integration strategy of \ourmethod allows it to automatically learn and exploit the most informative malicious signal, thereby effectively addressing evolving camouflaging strategies.
\ding{184}~The scalability of existing unsupervised GFD methods is limited by out-of-memory (OOM) issues when applied to large-scale real-world datasets such as AmazonVideo and YelpChi. In contrast, \ourmethod demonstrates better scalability, underscoring its effectiveness in safeguarding large-scale real-world graph applications like e-commerce and social networks. 
\ding{185}~Although existing TAGAD incorporate large language models to enhance feature alignment~\cite{xu2025text} or to generate auxiliary evidence~\cite{xu2025court}, they achieve limited performance on the TAGFD task. This is because malicious intent is often concealed by semantic camouflage employed by fraudsters, making it substantially harder to identify than anomalies in constructed benchmark datasets~\cite{ma2021comprehensive}.

\subsubsection{Ablation Study }
To examine the contribution of key design in \ourmethod, we conduct ablation studies on critical model components.

\noindent \textbf{Textual Feature Encoder.} We replace the LLM-based feature encoder with bag-of-words (BoW) and SentenceBERT representations \cite{reimers2019sentence} to assess the impact of semantic representation quality. As shown in Table \ref{table:Ablation:Feature}, the LLM-based encoder consistently achieves the highest detection performance. In contrast, although bag-of-words representations are the most commonly-adopted choice in GAD \cite{pan2023prem}, they fail to capture the critical semantic information required for TAGFD. These findings confirm that rich semantic embeddings are critical for detecting camouflaged fraudsters, aligning with our method’s emphasis on preserving subtle textual cues in an unsupervised setting. 

\begin{figure}[t]
\centering
    \includegraphics[width=\linewidth]{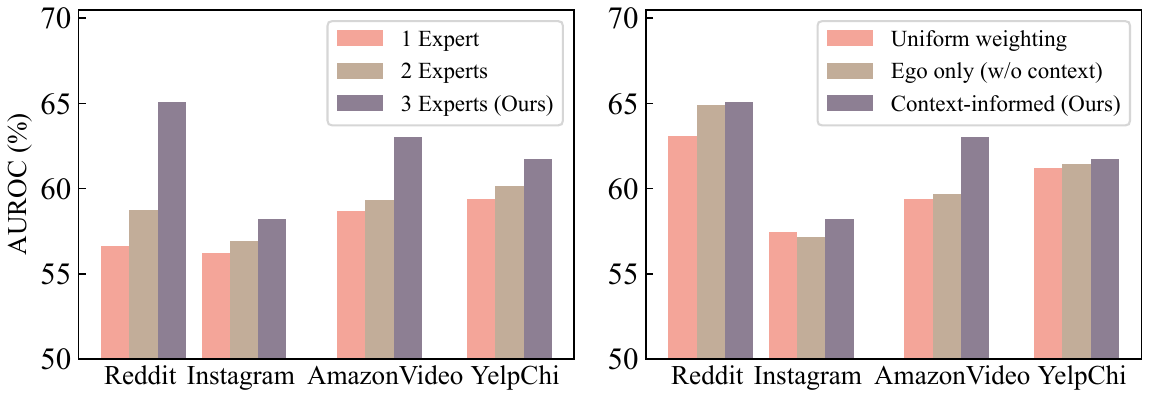}
\caption{Ablation on (L): \#experts and (R): gating mechanism.}
\label{fig:AblationMoE}
\end{figure}

\input{Tables/Ablation_Experts}
\noindent \textbf{Specialized Experts.} To quantify the contribution of each expert, we compare the full model with variants using only one or two experts. Table \ref{table:Ablation:Experts} shows that the full MoE achieves the best overall performance, and Figure~\ref{fig:AblationMoE} (left) demonstrates that adding more experts consistently improves the results. Together, these results demonstrate that successful TAGFD demands diverse fraud-indicative cues. Among the experts, the graph expert contributes most significantly, highlighting the importance of local structural deviations.

\noindent \textbf{Context-informed Gating.}  We evaluate the gating model by replacing it with two alternatives: uniform weighting and ego-only gating. As shown in Figure \ref{fig:AblationMoE} (right), incorporating local neighborhood context consistently improves performance across all datasets, highlighting the importance of incorporating community-specific information for weighting experts.  In contrast, uniform weighting leads to overall inferior results. Together, the experiment findings confirm that adaptively integrating fraud-indicative cues is critical for identifying evolved, camouflaged fraudsters.

\begin{figure}[t]
\centering
    \includegraphics[width=0.9\linewidth]{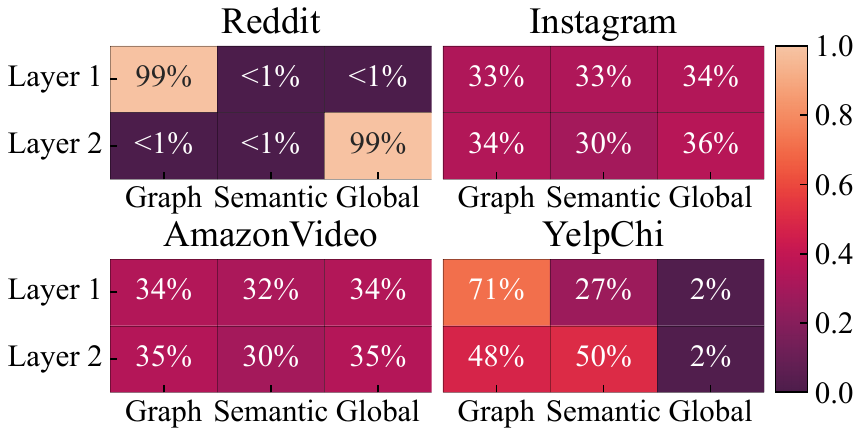}
\caption{Visualization of expert weight.}
\label{fig:CaseStudy:Weight}

\end{figure}

\begin{figure}[t]
\centering
    \includegraphics[width=0.9\linewidth]{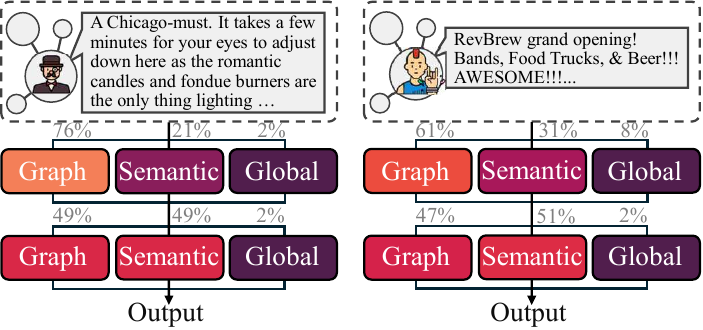}

\caption{Case study on YelpChi. }
\label{fig:CaseStudy:Yelp}
\end{figure}

\subsubsection{Visualization of Expert Allocation} 
To investigate how \ourmethod adaptively integrates fraud-indicative cues across different scenarios, we visualize the gating weights at the dataset and case levels.

\noindent \textbf{Dataset-level visualization} is shown in Figure~\ref{fig:CaseStudy:Weight}. 
On Instagram and AmazonVideo, the three experts receive relatively balanced weights. However, as illustrated in Figure~\ref{fig:AblationMoE}, replacing them with uniform weights results in a significant performance drop, which shows that the gating model also provides sample-specific fraud-indicative cues integration. 
On YelpChi and Reddit, the model demonstrates sparse activation. Specifically, YelpChi primarily relies on graph and semantic cues, while Reddit emphasizes the graph expert at layer 1 and the global expert at layer 2. This visualization result also shows that \ourmethod performs hierarchical cue integration through expert specialization at different layers. 

\noindent \textbf{Case Study on YelpChi.} To further investigate the gating model, we analyze the activation weights on YelpChi in detail. As illustrated in Figure~\ref{fig:CaseStudy:Yelp}, on the first MoE layer, different types of food venues exhibit distinct expert weight patterns. For luxury restaurants, the gating model assigns a higher weight to the graph expert compared to the weight reported in Figure ~\ref{fig:CaseStudy:Weight}, emphasizing the importance of reputation. In contrast, beer gardens rely more on subjective reviews,  resulting in relatively higher weights for the semantic expert. These observations confirm that \ourmethod can achieve adaptive integration of fraud-indicative cues.

To sum up, these observations confirm that \ourmethod can achieve adaptive integration of fraud-indicative cues. This flexibility not only enhances detection of evolved fraudsters in diverse real-world scenarios but also enables data-specific integration within the same datasets, leading to superior unsupervised TAGFD performance.

%% file: Tables/Ablation_Feature.tex
\begin{table}[t]
\centering
\resizebox{1.0\columnwidth}{!}{
\begin{tabular}{l|cccc}
\toprule
Method & Reddit & Instagram & AmazonVideo & YelpChi \\
\midrule
OpenAI (Ours) & \textbf{65.09±0.04} & \textbf{58.21±0.29} & \textbf{63.05±0.60} & \textbf{61.74±0.04} \\
\midrule
SentenceBERT & 61.45±0.10 & 53.23±1.67 & 53.95±3.82 & 53.79±0.16 \\
BoW & 48.88±0.04 & 51.52±0.08 & 51.58±0.05 & 39.93±0.04 \\
\bottomrule
\end{tabular}
}
\caption{Ablation study on text encoder. Result in AUROC (\%). }
\label{table:Ablation:Feature}

\end{table}

%% file: Tables/Ablation_Experts.tex




\begin{table}[t]
\centering
\resizebox{1.0\columnwidth}{!}{
\begin{tabular}{ccc|cccc}
\toprule
\multicolumn{3}{c|}{Experts} & \multicolumn{4}{c}{AUROC (\%)} \\
Graph & Global & Semantic & Reddit & Instagram & AmazonVideo & YelpChi \\
\midrule
\ding{51} & \ding{51} & \ding{51} & \textbf{65.09$\pm$0.04} & \textbf{58.21$\pm$0.29} & \textbf{63.05$\pm$0.60} & \textbf{61.74$\pm$0.04} \\
\midrule
--     & \ding{51} & \ding{51} & 55.74$\pm$0.51 & 56.55$\pm$1.55 & 58.97$\pm$3.37 & 60.32$\pm$0.17 \\
\ding{51} & --     & \ding{51} & 59.96$\pm$1.50 & 56.52$\pm$0.57 & 57.69$\pm$5.59 & 60.24$\pm$0.18 \\
\ding{51} & \ding{51} & --     & 60.46$\pm$1.71 & 57.65$\pm$0.34 & 61.35$\pm$1.38 & 59.90$\pm$0.07 \\

\ding{51} & --     & --     & 61.40$\pm$0.34 & 57.95$\pm$0.47 & 61.35$\pm$1.56 & 60.29$\pm$0.06 \\
--     & \ding{51} & --     & 54.00$\pm$0.33 & 57.06$\pm$0.35 & 60.00$\pm$0.53 & 58.14$\pm$0.35 \\
--     & --     & \ding{51} & 54.41$\pm$0.29 & 53.55$\pm$1.98 & 54.64$\pm$7.80 & 59.74$\pm$0.22 \\
\bottomrule
\end{tabular}
}
\caption{Ablation study on expert participation.}
\label{table:Ablation:Experts}
\end{table}

%% file: Appendix_Content.tex
\input{Chapters/Appendix_RW}

\input{Tables/Experiments_Datasets}
\input{Chapters/Appendix_Theory}

\section{Algorithm Description}
The procedure of training and inference \ourmethod is summarized in Algorithm~\ref{alg:algorithm}. 

\begin{algorithm}[tb]
\caption{\ourmethod}
\label{alg:algorithm}
\textbf{Input}: Text-Attributed graph $G=(\mathcal{V}, \mathcal{E}, \mathcal{T})$, LLM: pretrained and frozen text encoder, $E$: Training epochs, lr: Learning rate, $\alpha$: weight for $\mathcal{L}_\text{gating}$,  $\beta$:  weight for $\mathcal{L}_\text{OC}$  \\
\textbf{Output}: Anomaly scores $\mathbf{s}=\{s_1, ..., s_n \}$ \\
\begin{algorithmic}[1] 
\STATE Randomly initialize parameters of the ego-decoupled MoE layers $f=f^{[1]} \circ f^{[2]} \circ ...\circ  f^{[L]}$, where $f^{[l]}$ includes experts $e_k^{[l]}$  and the gating model $g^{[l]}_k$, where $k\in \{\text{graph},\text{semantic},\text{global} \}$. 
\STATE $\mathbf{H}^{[0]} \leftarrow \text{LLM}(\mathcal{T})$
\STATE // \textit{Training phase}
\FOR{$epoch=1,...,E$}
    \STATE $\mathcal{L}_{\text{expert}} \leftarrow 0 $
    \STATE $\mathcal{L}_{\text{gating}} \leftarrow 0 $
    \FOR{$l=1,...,L$}
        \STATE  $\mathbf{H}^{[l]}  \leftarrow f^{[l]}(\mathbf{H}^{[l-1]}, \textbf{A})$ 
        \STATE $\mathcal{L}_{\text{expert}}  \leftarrow \mathcal{L}_{\text{expert}} + \sum_{k} \frac{1}{N} \sum_{i=1}^N \| e_k^{[l]}(\mathbf{h}_i^{[l-1]}, \mathbf{A}) \|_2^2$
        \STATE $\mathcal{L}_{\text{gating}}  \leftarrow \mathcal{L}_{\text{gating}} +\sum_{k} - \frac{1}{N} \sum_{i=1}^N  g_{i,k}^{[l]} \log (g_{i,k}^{[l]} + \epsilon)$
    \ENDFOR
    \STATE Compute $\mathbf{s}$, where $s_i = \sigma(\|\mathbf{h}^{\text{L}}_i\|_2)$
    \STATE $\mathcal{L}_\text{OC} = \frac{1}{N} \sum_{i=1}^N \text{BCE}(s_i, 0)$
    \STATE $\mathcal{L} = \mathcal{L}_\text{expert} + \alpha \mathcal{L}_\text{gating} + \beta \mathcal{L}_\text{OC}$
    \STATE Update model $f$ by back-propagating $\mathcal{L}$ with learning rate lr.
\ENDFOR
\STATE // \textit{Inference phase}
\STATE $ \mathbf{H}^{[L]} \leftarrow f(\mathbf{H}^{[0]}, \textbf{A})$
\STATE Compute $\mathbf{s}$, where $s_i = \sigma(\|\mathbf{h}^{\text{L}}_i\|_2)$
\STATE \textbf{return} $\mathbf{s}$
\end{algorithmic}
\end{algorithm}

\section{Complexity Analysis}
We discuss the time complexity of each component in \ourmethod. Let $n$ and $m$ denote the number of nodes and edges in the input graph, respectively. The textual encoder maps raw text attributes to node embeddings with a cost of $\mathcal{O}(n)$. In each training epoch, the GNN in the graph expert costs  $\mathcal{O}(m)$ to perform message passing, while the semantic expert and global expert have a complexity of $\mathcal{O}(n)$. The context-informed gating model computes expert weights based on ego and neighborhood embeddings, which requires an additional $\mathcal{O}(m + n)$ cost for neighborhood aggregation and linear projection. As the computation of anomaly score and loss operates in $\mathcal{O}(n)$ complexity, the overall per-epoch training complexity is $\mathcal{O}(L(m + n))$, where $L$ denotes the number of MoE layers. Overall, \ourmethod scales linearly with the number of nodes and edges, making it suitable for real-world large-scale TAGFD scenarios.

\input{Chapters/Appendix_Setting}
\input{Chapters/Appendix_Results}

%% file: Chapters/Appendix_RW.tex
\section{Related work in details}
\subsection{Fraud Detection on Attributed Graph}
Graph fraud detection (GFD) aims to identify fraudulent activities in real-world graph applications, such as financial fraud~\cite{li2022internet}, fake reviews~\cite{yu2023group}, and spamming~\cite{hu2023cost}. Earlier works treat GFD as a class-imbalance classification problem and incorporate techniques such as sampling~\cite{liu2021pick}. However, camouflaged fraudsters bring unique challenges, as fraudsters can blend into benign communities in graph structure through heterophily edges~\cite{gao2023addressing}. As a result, later studies explicitly incorporate this knowledge into model design. For example, GHRN~\cite{gao2023addressing} directly prunes heterophily edges, while BWGNN~\cite{tang2022rethinking} employs spectral GNNs to reduce the 'right-shift’ phenomenon caused by fraudsters. 

Despite their promising results, the reliance on labels restricts their applicability in unsupervised scenarios, which motivates early attempts to leverage graph anomaly detection (GAD) by treating fraudsters as outliers in graphs.  To address this problem, early attempts have tried to utilize graph anomaly detection (GAD) methods to treat fraudsters as outliers in the graph. These methods can be broadly categorized into three paradigms: contrastive learning~\cite{liu2021anomaly}, reconstruction~\cite{ding2019deep}, and affinity-guided methods~\cite{pan2023prem}. However, the camouflaged fraudsters can disrupt unsupervised learning objectives, resulting in issues such as anomaly overfitting and homophily traps~\cite{he2024ada}. Therefore, follow-up unsupervised GFD studies utilize heuristic assumptions to address camouflaged fraudsters. For example, TAM~\cite{qiao2023truncated} and ADA-GAD~\cite{he2024ada} prune heterophily edges by iteratively computing pseudo-labels, while HUGE~\cite{pan2025label} establishes a label-free heterophily measure to guide the fraud detection.

\subsection{Fraud Detection on Text-Attributed Graph}
As many real-world graph applications are inherently associated with rich textual context, text-attributed GFD (TAGFD) has emerged as an important research direction. However, beyond structure-level blending, evolved fraudsters introduce new challenges through semantic camouflage, deliberately concealing malicious intent by imitating text generated by benign users. To address this, it is necessary to effectively utilize textual information. For example, LESS4FD~\cite{ma2024fake} constructs heterogeneous graphs with rich semantic information to provide strong support for fake news detection. With the advancement of pretrained foundation models,  DGB~\cite{li2025dgp} and GuARD~\cite{pang2025guard} finetune a large language model(LLMs) to enable joint understanding of fraud patterns and graph structures. To improve efficiency, FLAG alternatively uses LLM as teachers to guide smaller detection models in learning discriminative patterns~\cite{yang2025flag}.

However, in real-world applications, annotating semantically camouflaged fraudsters requires substantial manual effort, making unsupervised TAGFD methods increasingly important given rapidly expanding use cases. Despite their practical relevance, most existing attempts focus on artificial settings, namely text-attributed GAD benchmarks, and remain highly challenged by semantic camouflage. One line of work leverages the zero-shot capabilities of LLM. For instance, CoLL~\cite{xu2025court} incorporates LLMs to generate anomaly evidence from both textual content and graph structure to assist downstream anomaly detection. However, as evolved fraudsters deliberately mimic benign user behaviors, LLMs are often misled, resulting in incorrect or misleading evidence. Another line of methods integrates structural and textual information into unified graph contrastive learning frameworks~\cite{liu2025towards,xu2025text}. Nevertheless, semantic camouflage typically introduces only small ego-neighbor divergence compared to the injected structure and feature anomalies~\cite{ding2019deep}, 
which undermines the reliability of contrastive training objectives. These limitations motivate our work to  address semantic camouflage in unsupervised TAGFD, thereby safeguarding graph applications against evolving fraudsters.

%% file: Tables/Experiments_Datasets.tex
\begin{table}[t]
\centering
\begin{tabular}{lrrr}
\hline
Dataset & $|\mathcal{V}|$ & $|\mathcal{E}|$ & Avg. $L$ \\
\hline
Reddit     & 18{,}574 & 64{,}469    & 142 \\
Instagram  & 8{,}026  & 85{,}520    & 15  \\
Amazon     & 37{,}126 & 3{,}658{,}396 & 98  \\
YelpChi    & 67{,}395 & 16{,}553{,}904 & 139 \\
\hline
\end{tabular}
\caption{Dataset statistics.}
\label{tab:dataset_stats}
\end{table}

%% file: Chapters/Appendix_Theory.tex
\section{Analyzing Collapse Risk of Training Objectives}

While the rarity-driven training pipeline enables unsupervised learning in \ourmethod, a careful reader may note that the one-class (OC) loss could potentially encourage trivial solutions, i.e., assigning uniformly zero fraud scores to all instances. In the following, we show that, provided the model is not overly over-parameterized, such collapse does not occur. 

Assume the model has $L$ MoE layers. The final embedding of node $v_i$ can be expressed as

\[
\mathbf{h}_i^{\mathrm{final}} = \mathbf{h}_i^{[L-1]} + \sum_{k} g_{i,k}^{[L]}\, e_k^{[L]}(\mathbf{h}_i^{[L-1]}, \mathbf A), 
\]

and the OC loss is defined as
\[
\begin{aligned}
\mathcal{L}_{\mathrm{OC}} 
=& \frac{1}{N}\sum_{i=1}^N \mathrm{BCE}\big(\sigma(\|\mathbf{h}_i^{\mathrm{final}}\|_2), 0\big) \\
=& - \frac{1}{N}\sum_{i=1}^N \log\Big(1 - \sigma(\|\mathbf{h}_i^{\mathrm{final}}\|_2)\Big) \\
=& - \frac{1}{N}\sum_{i=1}^N \log\Big(1 -\\
  &\sigma(\|\mathbf{h}_i^{[L-1]} + \sum_{k} g_{i,k}^{[L]}\, e_k^{[L]}(\mathbf{h}_i^{[L-1]}, \mathbf A)\|_2)\Big).
\end{aligned}
\]

From this, it follows that minimizing $\mathcal{L}_{\mathrm{OC}}$ is equivalent to minimizing the $\ell_2$ norm of the gating-weighted expert residuals:

\[
\min \mathcal{L}_{\mathrm{OC}} \;\Longleftrightarrow\; 
\min \sum_i \Big\| \sum_{k} g_{i,k}^{[L]} e_k^{[L]}(\mathbf{h}_i^{[L-1]}, \mathbf A) \Big\|_2.
\]

Therefore, to analyze potential collapse, we need to examine both the gating model and the experts to see if they could potentially collapse.

\noindent\textbf{Statement 1: The gating model does not collapse.}  Although the interaction between the gating model and OC loss is complex, the gating mechanism employs a softmax function that ensures the expert weights sum to one. As a result, the gating model cannot trivially collapse to zero weight for all experts.

\noindent \textbf{Statement 2: Experts do not trivially collapse.}  
As both the expert loss and the OC loss encourage minimizing the output signal of experts, we analyze each expert using the expert loss to show why collapse is unlikely. Similar procedure can be applied to the OC loss.

\begin{itemize}
    \item \textbf{Graph expert}: As expert loss minimizes the Euclidean distance between ego features and aggregated neighbor features, it can be seen as reconstructing the neighborhood features using ego information~\cite{roy2024gad}.
    \item \textbf{Global expert}: The global expert is identical to the graph expert operating on a clique graph structure. Therefore, similar reconstruction statement holds.  
    \item \textbf{Semantic expert}: The semantic expert directly employs an autoencoder to model benign patterns. By approximating the manifold of input data with limited parameters, the encoder acts as a denoiser~\cite{kingma2013auto} to capture the distribution of normal semantic. 
\end{itemize}

In this sense, each expert is trained with an autoencoder-like loss that minimize the discrepancy to learn benign manifold, and return the discrepancy between ego and reconstructed features to highlight fraud-indicative cues. Therefore, despite the OC training objective, on an expert-level, it is as collapse-tolerant as an autoencoder-based anomaly detectors, which is widely utilized in anomaly and fraud detection tasks~\cite{ding2019deep,he2024ada}. 

\begin{figure}[t]
\centering
    \includegraphics[width=\linewidth]{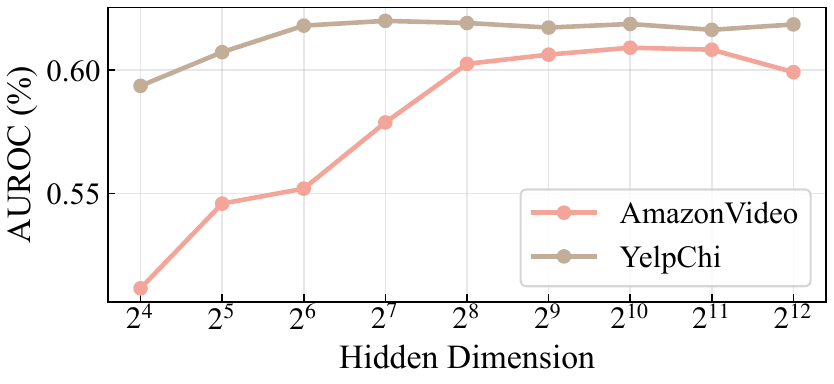}

\caption{Over-parameterization study on YelpChi and Amazon datasets.}

\label{fig:Ablation:Sensitivity}
\end{figure}

Despite this intuition, the risk of overfitting to residuals remains, as sufficiently wide neural networks can approximate arbitrary functions~\cite{cybenko1989approximation}.  This can give rise to issues such as anomaly overfitting and homophily traps~\cite{he2024ada} in GFD, thereby reducing performance. . To study the effect of over-parameterization, we employ a two-layer MLP as the encoder for each expert and investigate how varying the hidden dimension affects results. As shown in Figure~\ref{fig:Ablation:Sensitivity}, \ourmethod exhibits solid performance even when over-parameterized, demonstrating its robustness.

%% file: Chapters/Appendix_Setting.tex
\section{Dataset and Implementation Details}

\input{Tables/Experiments_Hyperparams}

Detailed statistics of the datasets are summarized in Table~\ref{tab:dataset_stats}. All experiments are conducted on a Windows desktop equipped with 32~GB RAM and an RTX~4090 GPU with 24~GB VRAM. For implementation details of \ourmethod, we stack two MoE layers, where each expert is implemented with a single corresponding encoder layer. The hyperparameter settings are reported in Table~\ref{tab:hyperparams}.

%% file: Tables/Experiments_Hyperparams.tex
\begin{table}[t]
\centering
\resizebox{1.0\columnwidth}{!}{
\begin{tabular}{lcccc}
\hline
Dataset & Reddit & Instagram & AmazonVideo & YelpChi \\
\hline
Epoch  & 1200 & 15 & 15 & 450 \\
$\alpha$ & 5.0 & 10.0 & 0.1 & 0.1 \\
$\beta$  & 0.1 & 10.0 & 1.0 & 10.0 \\
learning rate & 1e-3 & 5e-5 & 5e-5 & 1e-3 \\
\hline
\end{tabular}
}
\caption{Hyperparameter settings.}
\label{tab:hyperparams}

\end{table}

%% file: Chapters/Appendix_Results.tex
\input{Figures/Fig_GridSearch}

\section{Addition Experiments Results}

\subsection{Sensitivity}
To investigate the effect of the hyperparameters $\alpha$ and $\beta$, we perform a grid search for them. As illustrated in Figure~\ref{fig:gridsearch}, different datasets require different hyperparameters. The grid search results are shown in Figure~\ref{fig:gridsearch}. Overall, the performance varies smoothly across the search space, indicating that the proposed method is not overly sensitive to these hyperparameters. This robustness is particularly desirable in unsupervised TAGFD, where labeled data for validation purposes are unavailable. Hence, the observed stability further supports the practicality of \ourmethod in real-world scenarios with diverse and evolving fraud patterns.

%% file: Figures/Fig_GridSearch.tex
\begin{figure}[t]  
    \centering
    \begin{subfigure}[b]{0.22\textwidth}  
        \centering
        \includegraphics[width=\linewidth]{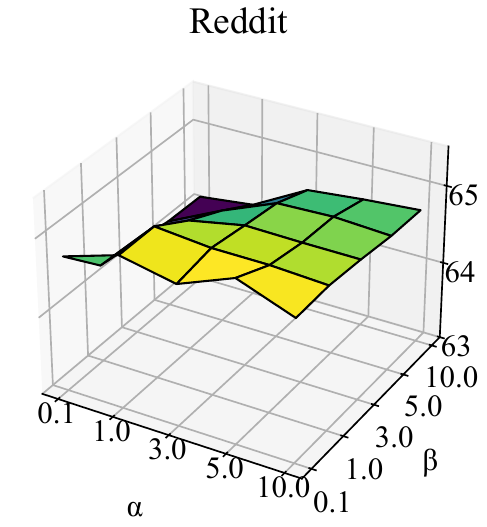}
    \end{subfigure}
    \hfill
    \begin{subfigure}[b]{0.22\textwidth}
        \centering
        \includegraphics[width=\linewidth]{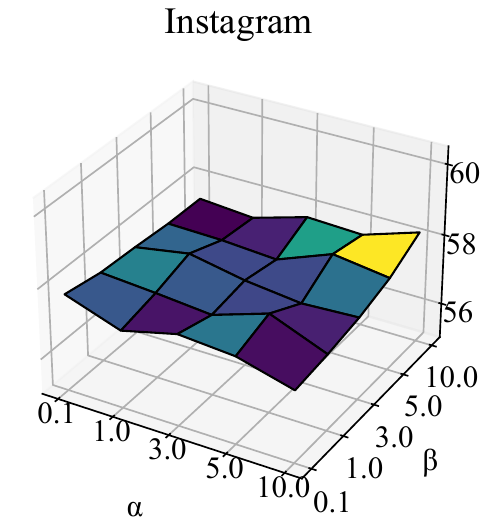}

    \end{subfigure}
    \hfill
    \begin{subfigure}[b]{0.22\textwidth}
        \centering
        \includegraphics[width=\linewidth]{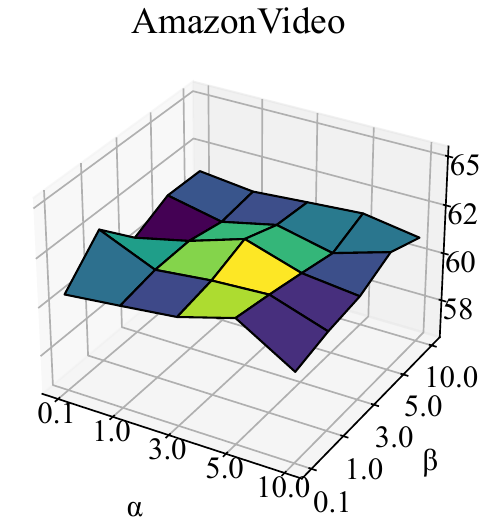}

    \end{subfigure}
    \hfill
    \begin{subfigure}[b]{0.22\textwidth}
        \centering
        \includegraphics[width=\linewidth]{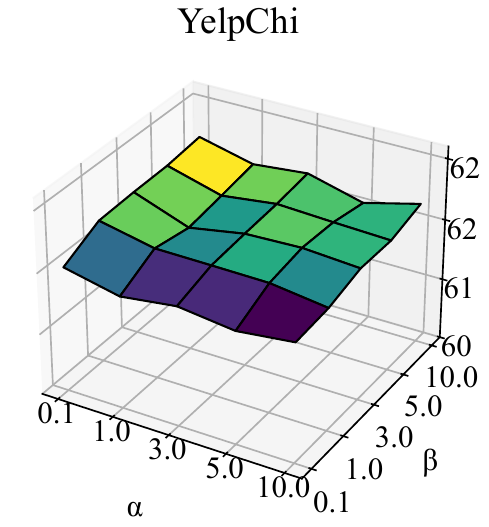}
    \end{subfigure}

    \caption{Grid search results for $\alpha$ vs. $\beta$ (surface height indicates AUROC).}
    \label{fig:gridsearch}
\end{figure}